\documentclass[letterpaper, 10 pt, journal, twoside]{IEEEtran}
\usepackage{amsmath}

\newcommand{\vect}[1]{\boldsymbol{#1}}
\newcommand{\mat}[1]{\boldsymbol{#1}}

\newcommand{\diffs}[3]{\frac{\partial^2 #1}{
\ifx#2#3 
\partial #2^2
\else
\partial #2 \partial #3
\fi
}}

\newcommand{\cv}{\vect{c}}

\newcommand{\fv}{\vect{f}}
\newcommand{\gv}{\vect{g}}

\newcommand{\pv}{\vect{p}}

\newcommand{\rv}{{\vect{r}}}

\newcommand{\uv}{\vect{u}}
\newcommand{\vv}{\vect{v}}

\newcommand{\xv}{\vect{x}}

\newcommand{\zv}{\vect{z}}

\newcommand{\omegav}{\vect{\omega}}

\newcommand{\IIm}{\mat{I}}

\newcommand{\Em}{\mat{E}}

\newcommand{\Qm}{\mat{Q}}

\newcommand{\Phim}{\mat{\Phi}}

\usepackage{amsmath,amsfonts}
\usepackage{algorithm, algpseudocode}
\usepackage{array}
\usepackage[caption=false,font=normalsize,labelfont=sf,textfont=sf]{subfig}
\usepackage{textcomp}
\usepackage{stfloats}
\usepackage{url}
\usepackage{verbatim}
\usepackage{graphicx}
\usepackage{cite}
\usepackage{xcolor}
\usepackage{enumitem}

\usepackage{soul}
\usepackage{microtype}

\usepackage{xcolor}
\definecolor{lgray}{gray}{0.30}
\usepackage{eso-pic}
\usepackage{hyperref}

\AddToShipoutPictureBG*{%
  \AtPageUpperLeft{%
    \setlength{\unitlength}{1mm}%
    \put(-18.5,-9){\makebox(\paperwidth,0)[c]{\parbox{0.8\textwidth}{IEEE Robotics and Automation Letters, 2025, vol. 10, no. 2, pp. 1265-1272}}}
  }
}

\AddToShipoutPictureBG*{%
  \AtPageUpperLeft{%
    \setlength{\unitlength}{1mm}%
    \put(-9.5,-14){\makebox(\paperwidth,0)[c]{\parbox{0.9\textwidth}{DOI: \href{https://ieeexplore.ieee.org/document/10806593}{10.1109/LRA.2024.3519908}}}}
  }
}

\newcommand{\revise}{\textcolor{black}}
\begin{document}

\title{Non-Gaited Legged Locomotion with Monte-Carlo Tree Search and Supervised Learning}

\author{Ilyass Taouil\textsuperscript{1,2,3,*}, Lorenzo Amatucci\textsuperscript{1,*}, Majid Khadiv\textsuperscript{3}, Angela Dai\textsuperscript{2}, Victor Barasuol\textsuperscript{1}, \\Giulio Turrisi\textsuperscript{1}, Claudio Semini\textsuperscript{1} 
\thanks{This
work was supported by and in collaboration with INAIL, project "Sistemi Cibernetici
Collaborativi - Robot Teleoperativo".}
\thanks{This paper was recommended for publication by Associate Editor Brian Plancher and Editor
Abderrahmane Kheddar upon evaluation of the Reviewers’ comments. \textsuperscript{*} Equal Contribution} 
\thanks{\textsuperscript{1} Dynamic Legged Systems Laboratory, Istituto Italiano di Tecnologia (IIT), Genova, Italy. E-mail: name.lastname@iit.it}
\thanks{\textsuperscript{2} 3D AI Laboratory, Technical University of Munich (TUM), Germany. E-mail: name.lastname@tum.de}
\thanks{\textsuperscript{3} ATARI Laboratory, MIRMI, Technical University of Munich (TUM), Germany. E-mail: name.lastname@tum.de}
}


\maketitle

\begin{abstract}
Legged robots are able to navigate complex terrains by continuously interacting with the environment through careful selection of contact sequences and timings. However, the combinatorial nature behind contact planning hinders the applicability of such optimization problems on hardware. In this work, we present a novel approach that optimizes gait sequences and respective timings for legged robots in the context of optimization-based controllers through the use of sampling-based methods and supervised learning techniques. We propose to bootstrap the search by learning an optimal value function in order to speed-up the gait planning procedure making it applicable in real-time. To validate our proposed method, we showcase its performance both in simulation and on hardware using a 22 kg electric quadruped robot. The method is assessed on different terrains, under external perturbations, and in comparison to a standard control approach where the gait sequence is fixed a priori.
\end{abstract}

\vspace{-5pt}
\section{Introduction}
\label{sec:introduction}

\IEEEPARstart{L}{egged} robots traverse the world by making and breaking contacts with the environment. In doing so, they need to decide over a set of discrete and continuous optimization variables, e.g., the sequence of end-effectors establishing contact (discrete) and the contact timings and forces (continuous). Classically, the switching nature of the problem has been relaxed to cast the motion planning problem as a general nonlinear program (NLP) \cite{tassa2012synthesis},\cite{mordatch2012discovery},\cite{winkler2018gait}. However, these approaches are prone to poor local minima and they usually need a good warm-start to yield physically plausible trajectories. Hence, they have shown little success in the real world \cite{koenemann2015whole}, \cite{neunert2018whole}. Recent works have focused on handling contact as an explicit phenomenon formulating an optimization problem with a mixture of continuous and discrete optimization variables. While recent advances have enabled solving the continuous optimization for a given contact sequence in a receding horizon fashion \cite{meduri2023biconmp}, \cite{mastalli2023agile}, \cite{grandia2023perceptive}, solving the hybrid problem online is still out of reach.

One interesting approach to solve the hybrid optimization problem is to cast it as a Mixed-Integer Program (MIP) \cite{deits2014footstep} for kinematically feasible footstep planning on uneven terrain. Several extensions of this algorithm were later proposed to solve for the gait sequence \cite{aceituno2017simultaneous}, or to include the robot dynamics \cite{ponton2021efficient}. All of these works used a convex relaxation of the dynamics and environment geometry to make a fast resolution of the problem tractable. Later works have tapped into relaxing the integer optimization to an L1 norm minimization problem \cite{tonneau2020sl1m} and managed to solve the problem of contact patch selection in real-time for quadrupedal locomotion \cite{corberes2023perceptive}. However, in the general case of nonlinear dynamics and complex geometry of the environment, solving a non-convex MIP is an NP-hard problem and existing solvers do not scale well with the number of discrete decision variables. 

In \cite{Bratta-ContactNet}, the authors demonstrated the potential of using supervised learning to determine the contact scheduling and positioning for a quadrupedal robot. Recent advancements in Reinforcement Learning (RL) offered a novel point of view on the timing problem. RL-based controllers do not explicitly tackle the problem of contact timing. Instead, during the learning process, they automatically discover behaviors that involve changes in contacts embedding them in the network representation \cite{Miki-RL}. Nevertheless, most RL-based controllers incorporate a gait timer as an input to the network, which imposes a bias on the selected contact pattern \cite{Hwangbo-RL}. While these approaches have shown incredible performance on real hardware, \revise{heavy reward engineering is required to generate new motions and an ad-hoc domain randomization procedure is required for successful sim-to-real.}

\revise{As reviewed above, MIP-based approaches efficiently use the knowledge about dynamics, geometries, and constraints of both the robot and the environment to generate robot motions in various scenarios. Although uncertainties can be taken into account to achieve a robust MIP-based controller, generating complex movements is not straightforward. RL-based approaches, instead, can achieve complex behaviors to tackle very challenging scenarios. However, they do not take into consideration the knowledge about the structure of the problem and rely on sampling in simulation to find optimal control policies. 
To benefit from the power of each method, a framework that can efficiently use the structure of the problem and use machine learning to reduce the online computation is required. One approach is to use continuous RL algorithms for contact planning \cite{tsounis2020deepgait,li2023hierarchical}. However, these approaches ignore this structure in which the contact planning problem is mostly a decision-making process over \emph{discrete} variables.}


Monte-Carlo Tree Search (MCTS) has recently emerged and shown promise as an alternative to solve the hybrid optimization problem for locomotion \cite{amatucci_mcts} and object manipulation \cite{zhu2023efficient}. Both of these works have shown that MCTS can dramatically reduce the computation time in comparison with MIP, while slightly compromising the optimality of the solution. Recently, \cite{dhedin2024diffusion,akizhanov2024learning} used a MCTS formulation to select which surface to step onto, among all steppable surfaces. However, none of these works have managed to run MCTS in real-time on real hardware to adapt the discrete optimization variables reactively.

In this work, we propose a novel approach that builds on \cite{amatucci_mcts} that leverages MCTS to optimize the gait sequence and timings for quadrupedal locomotion, combining it with offline learning \cite{bertsekas2024model}, \cite{warm_start_memmo}, \cite{safesteps} to make it applicable in real-time. In particular, our core contributions are the following:

\begin{itemize}
    \item We introduce a simple and effective learning-based strategy to enhance the real-time capability of an MCTS algorithm for the purpose of non-gaited locomotion.
    \item We carry out an extensive analysis of the MCTS parameters and their influence on the robot's performance.
    \item We demonstrate, to the best of our knowledge, the first-ever successful real-time implementation of a sampling-based method for non-gaited locomotion on a real quadruped robot.
\end{itemize}

The rest of the paper is organized as follows: Sec.~\ref{sec:vanilla_mcts} introduces the vanilla MCTS algorithm, providing a foundation and context for our research. Sec.~\ref{sec:mcts_parametrization} presents a detailed and extensive analysis of the MCTS parameters analyzing their impact on the robot's performance. In Sec.~\ref{sec:mcts_speedup}, we describe how supervised learning can be used in combination with MCTS to enable real-time execution in the real world. Section~\ref{sec:results} presents the simulation and experimental results, showcasing the approach on an electric quadruped robot. Finally, Sec.~\ref{sec:conclusions} concludes the paper, summarizing our findings and suggesting directions for future research.

\begin{figure}[t]
    \centering
    \includegraphics[width=0.9\linewidth]{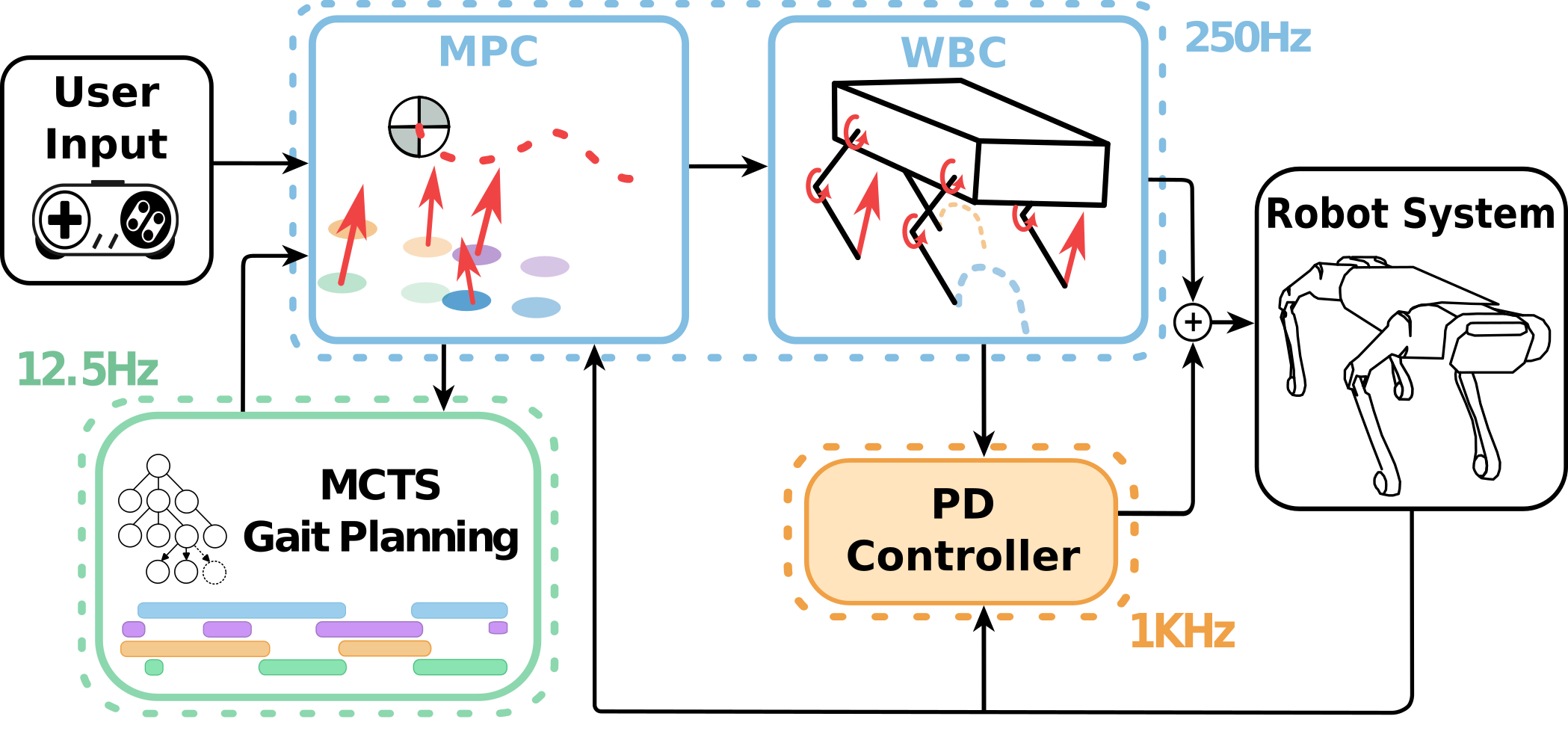}
    \caption{Control block diagram of the proposed approach based on the previous work \cite{amatucci_mcts}. The green block is executed at $12.5$ Hz, the blue block at $250$ Hz, and the orange block at $1K$ Hz.}
    \vspace{-15pt}
    \label{fig:block_scheme}
\end{figure}
\vspace{-5pt}
\section{Gait Planning using MCTS}
\label{sec:vanilla_mcts}

In this section, we present the MCTS gait planning architecture and describe how the MCTS iterative steps are adapted in the context of contact planning for controlling a multi-legged system. For the remainder of this paper, when referring to the terms \textit{gait planning} or \textit{contact planning} we mean in both occasions the optimization of both the contact sequence and respective contact timings of the end-effectors.

The proposed pipeline is shown in Fig.~\ref{fig:block_scheme}. A Model Predictive Control (MPC) framework receives, as input, a reference velocity from the user and an optimized gait sequence from the MCTS gait planner. The input velocity is used to generate the reference trajectory for the robot's base, while the gait sequence is used to generate the reference stepping locations and regulate the respective contact timings. The MPC and Whole-Body Control (WBC) are then responsible for producing the desired joint torques to track the references.

The gait planning problem is formulated as a Markov Decision Process (MDP). The MDP state \textbf{s} describes the contact state of the system and it includes each end-effector contact status, swing time, and stance time. The contact status is defined as a binary value identifying if the end-effector is in contact or not. The swing and stance times are continuous variables that identify for how long the end-effector has been out-of-contact (i.e. in swing) or in contact (i.e. in stance). The action \textbf{a}, causing a state transition $\textbf{s' = f(s, a)}$, selects the next contact status among the feasible contacts (Sec.~\ref{sec:expansion}), thereby also modifying the respective swing or stance timings. Each state is assigned a prediction cost $P(s)$ specifying the expected cost to go if such a state is selected.

\begin{figure}[t]
    \centering
    \captionsetup{justification=centering}
    \includegraphics[width=0.9\linewidth]{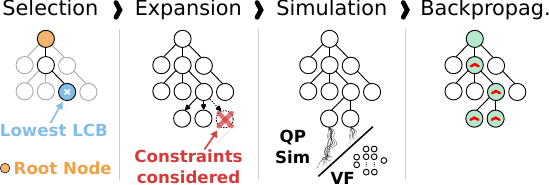}
    \caption{ \revise{MCTS search process \cite{amatucci_mcts} augmented with learned value function evaluation. \textit{Selection}: starting from the root node, a tree traversal is executed to find the node with the lowest cost. \textit{Expansion}: the selected node's children that respect the MCTS constraints are added to the search tree. \textit{Simulation}: the expanded nodes are assigned a prediction cost by solving several optimization problems and/or evaluating a learned value function network. \textit{Backpropagation}: the assigned prediction costs are backpropagated recursively to update the node's ancestors' costs.}}
    \vspace{-15pt}
    \label{fig:mcts_search_logic}
\end{figure}

The MCTS algorithm is used to solve the MDP and to optimize the gait sequence. Starting from a root node, describing the current contact state of the system, a search tree is created where each child node corresponds to a contact choice. At each iteration, the tree grows and deeper nodes in the tree constitute different contact plans up to a specified time horizon. \revise{ The MCTS search process is shown in Fig.~\ref{fig:mcts_search_logic} and is briefly described in the remainder of this section.} The MCTS algorithm terminates upon convergence, which is reached when a terminal node is selected during the MCTS \textit{selection} process. A node $n$ is terminal if its depth matches the MCTS planning horizon.


\subsection{Selection}
\label{sec:selection}

In the selection step, MCTS chooses which node to expand and explore next. This is a crucial part of the iterative process and is subject to the \textit{exploitation-exploration trade-off}. More precisely, when deciding the next expansion direction the process should balance between favoring the most promising node for a faster convergence or exploring other possibilities. \revise{In this work, as in \cite{amatucci_mcts}, we select the node to expand next based on the lowest Lower Confidence Bound (LCB) which incentivizes the exploration of nodes that are less visited in the tree. More precisely, the more a node is simulated, the smaller the difference between the LCB score and the node's average prediction cost becomes, where the latter is obtained by performing multiple MPC rollouts (Sec.~\ref{sec:simulation}). On the other hand, the fewer times a node is simulated, the higher the discount on the cost, incentivizing its selection for exploration.}

\subsection{Expansion}
\label{sec:expansion}
\revise{In the expansion step, the children of the selected node are added to the tree, each representing a potential contact state transition. For a legged system with $m$ legs, the number of children is $2^m$ since each leg can either be in contact or not. However, because we use a simplified SRB model in the MPC (Sec.~\ref{sec:simulation}), the leg dynamics are ignored, potentially leading to fast swing motions or short stance phases. To prevent this, we impose swing and stance time constraints during expansion. If a leg’s swing or stance time has not reached the minimum threshold, only child nodes that maintain the current swing or stance state are added, reducing the number of children. In this work, the minimum swing and stance times are set to 0.24 s and 0.16 s, respectively.}

\vspace{-2pt}
\subsection{Simulation \& Backpropagation}
\label{sec:simulation}
\revise{In the simulation step, each expanded node $n$ is assigned a prediction cost $\bar{P}_n$, which guides the MCTS search. 
To compute this cost, multiple MPC rollouts are solved for different gait sequences. For a given node $n_i$, we retrieve its contact sequence. If incomplete, we fill the sequence by sampling contacts from a uniform distribution that meets swing and stance constraints. This sampling is repeated multiple times to obtain a more reliable average cost. Each sequence is evaluated, in parallel, by solving an Optimal Control Problem (OCP) for a simplified SRB model, following the MPC formulation from \cite{turrisi2024sampling}. Our state and control vectors are defined as follows:}

\begin{align*}
    &\xv = [\pv_c, \dot\vv_c, \Phim, \boldsymbol{\omega}^b] \in R^{12}, \\
    &\uv = [\boldsymbol{f}_1, \boldsymbol{f}_2, \boldsymbol{f}_3, \boldsymbol{f}_4] \in R^{12}
    \label{state_equation}
\end{align*}

where $\pv_c \in R^3$ and $\dot\vv_c \in R^3$ are respectively the position and the linear velocity of the Center of Mass (CoM); $\Phim \in R^3$ is the base angular position (roll, pitch, and yaw); $\boldsymbol{\omega}^b \in R^3$ is the base angular velocity in the base frame, and $\boldsymbol{f}_i \in R^3$ the respective Ground Reaction Force (GRF) for the $i^{th}$ robot's foot. All the quantities, if not specified, are expressed in the world frame.

The cost of the OCP is defined as the combination of quadratic tracking and regularization terms, described as follows:
\revise{
\begin{equation}  
    \tilde{P} =\sum_{k = 0}^{N}\left(\|\boldsymbol{e}_{\boldsymbol{x}_{ k}}\|_{\boldsymbol{Q}_{\boldsymbol{x}}} + \|\boldsymbol{e}_{\boldsymbol{u}_{k}}\|_{\boldsymbol{R}_{u}} \right) 
h\label{eq::costOCP}
\end{equation}
}
where $\boldsymbol{e}_{(., k)}$ is the error with respect to the reference state and control at the $k^{th}$ prediction step, and $\Qm_{x}$ and $\boldsymbol{R}_u$ are diagonal weighting matrices.

Finally, the dynamic model is defined as 
\vspace{-2pt}
\begin{equation}  
    \hspace{-5pt}
    \resizebox{\columnwidth}{!}{
    
    $\left[\begin{array}{c}
    \dot{\pv}_{c} \\ 
    \dot{\vv}_{\mathrm{c}} \\ 
    \dot{\Phi} \\ 
    \dot{\mathcal{\omegav}^b}
    \end{array}\right]=
    \left[\begin{array}{c}
    \vv_{\mathrm{c}} \\ 
    1 / m \sum_{i=1}^4 \delta_i \fv_i+\gv \\ 
    \Em^{\prime-1}(\Phim) \omegav^b \\ 
    - \IIm_{\mathrm{c}}^{-1}\left( \omegav^b \times{ } \IIm_{\mathrm{c}}\right)  \omegav^b +\sum_{i=1}^4 \delta_{i} \IIm_{\mathrm{c}}^{-1}{ } \rv_{i} \times{ }\fv_i
    \end{array}\right]$}
    \label{eq:srbd_model}
\end{equation}

with $m$ characterizing the robot mass subjected to gravitational acceleration $\gv$ and $\IIm \in \mathbb{R}^{3 \times 3}$ the constant inertia tensor centered at the robot's CoM; $\Em^{\prime-1}$ is a mapping from the SRB angular velocity to Euler rates; and the displacement vector between the CoM position $\pv_c$ and the $i^{th}$ robot's foot $\pv_{f,i}$ is defined as $\rv_{i} = \pv_{f,i} - \pv_c \in \mathbb{R}^3$. Binary variables $\delta_i = \{0, 1\}$ indicate whether an end-effector makes contact with the environment, and can produce interaction forces, or not. These variables are extracted from the simulated gait sequence. 
\revise{
Finally, friction cone constraints are added to the optimization problem to limit the maximum and minimum GRF and avoid foot slippage. We define the QP problem to be solved starting from the cost function in \eqref{eq::costOCP} and imposing as equality constraint the discretized and linearized version of the dynamics in \eqref{eq:srbd_model}, while as inequality constraint we impose the outer pyramid approximation of the friction cones.} Once $\tilde{P}$ is computed by solving the QP, we update the prediction cost for each node $n$ that is simulated as follows:
\vspace{-3pt}
\begin{equation}
\bar{P}_n =  \frac{\sum_{m=1}^M  \bigg(\tilde{P}_{m} + \sum_{l=1}^L \lambda (T_{sw_r} - T_{sw, l})\bigg)}{M}
\label{eq:qps_cost}
\end{equation}

where $M$ is the number of times we solve the QP with different randomly completed contact sequences. We observed that by using only the cost $\tilde{P}$, the MCTS tends to choose in most cases the fastest allowed swing time. This behavior arises because the SRB model does not consider any component of the leg dynamics. To overcome this limitation, we included an additional discrete cost to drive each leg's swing time $T_{sw, l}$ towards a reference swing time $T_{sw_r}$. The constant $\lambda$ balances between faster and slower swing timings. Increasing the value of $\lambda$ makes the MCTS solution track the reference frequency making it less prone to show any contact timing adaptation in response to disturbances. As described in Sec.~\ref{sec:simulations_discretization_comparison}, the value $M$ is critical for the success of the algorithm. Due to its sampling-based nature, performing few simulations can lead to inaccurate prediction costs, whereas too many random sequences can lead to the impossibility of solving the MCTS algorithm within the replanning time budget.

\revise{In the backpropagation step, the prediction cost of each simulated node is propagated back to the respective parent node in a recursive fashion until the root node is reached. Propagating the cost back is important to improve the accuracy of the estimated cost for nodes closer to the root one.}


\section{MCTS Parametrization}
\label{sec:mcts_parametrization}

\begin{figure}[t]
    \centering
    \includegraphics[width=0.8\linewidth]{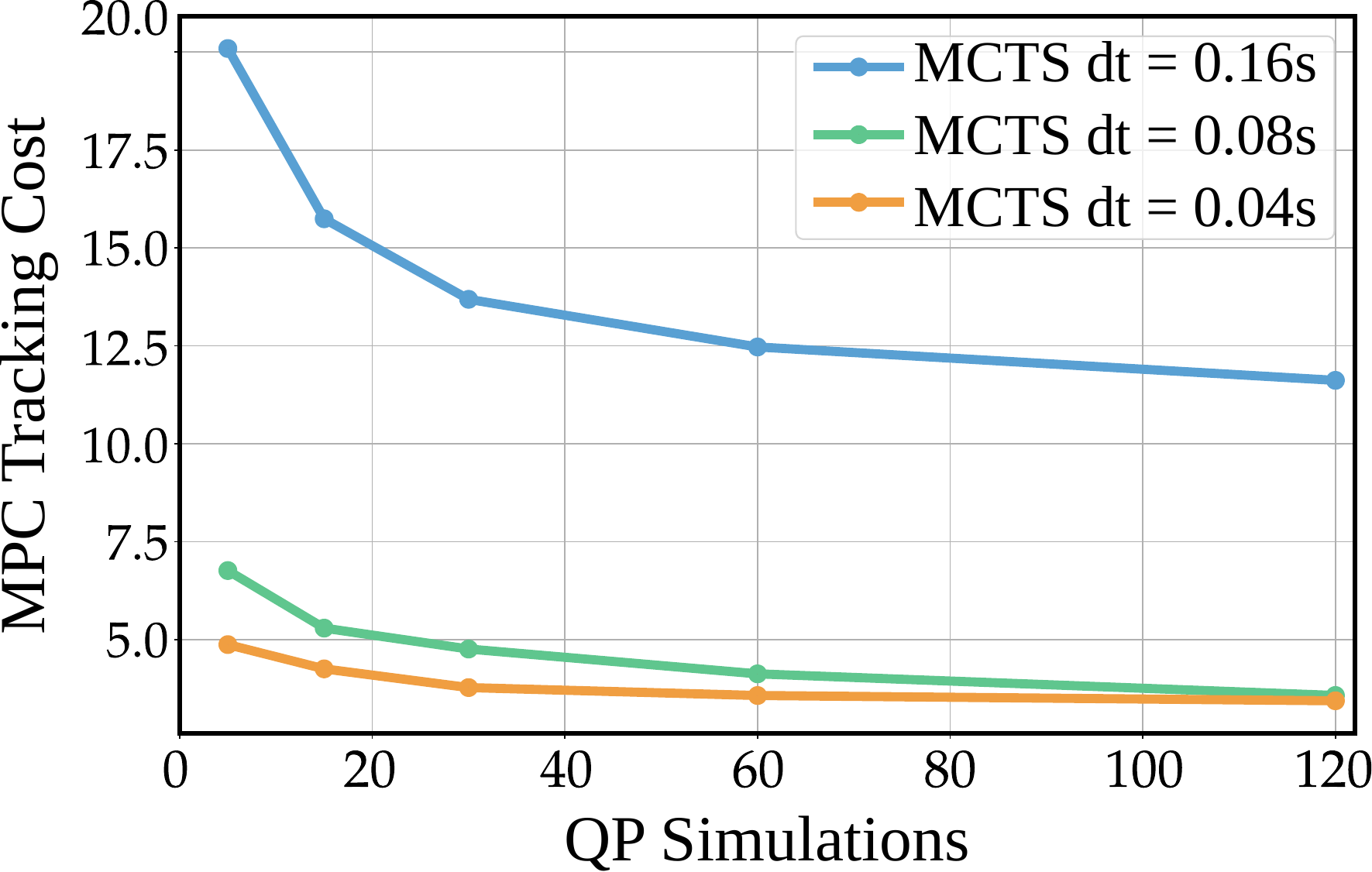}
    \caption{Comparison of the mean MPC tracking cost for different tree discretizations and increasing number of simulations while disturbing the system along different swing phases with a force of $150$ N for $100$ ms.}
    \vspace{-5pt}
    \label{fig:mcts_real_cost_comparison}
\end{figure}

\begin{figure}[t]
    \centering
    \includegraphics[width=0.8\linewidth]{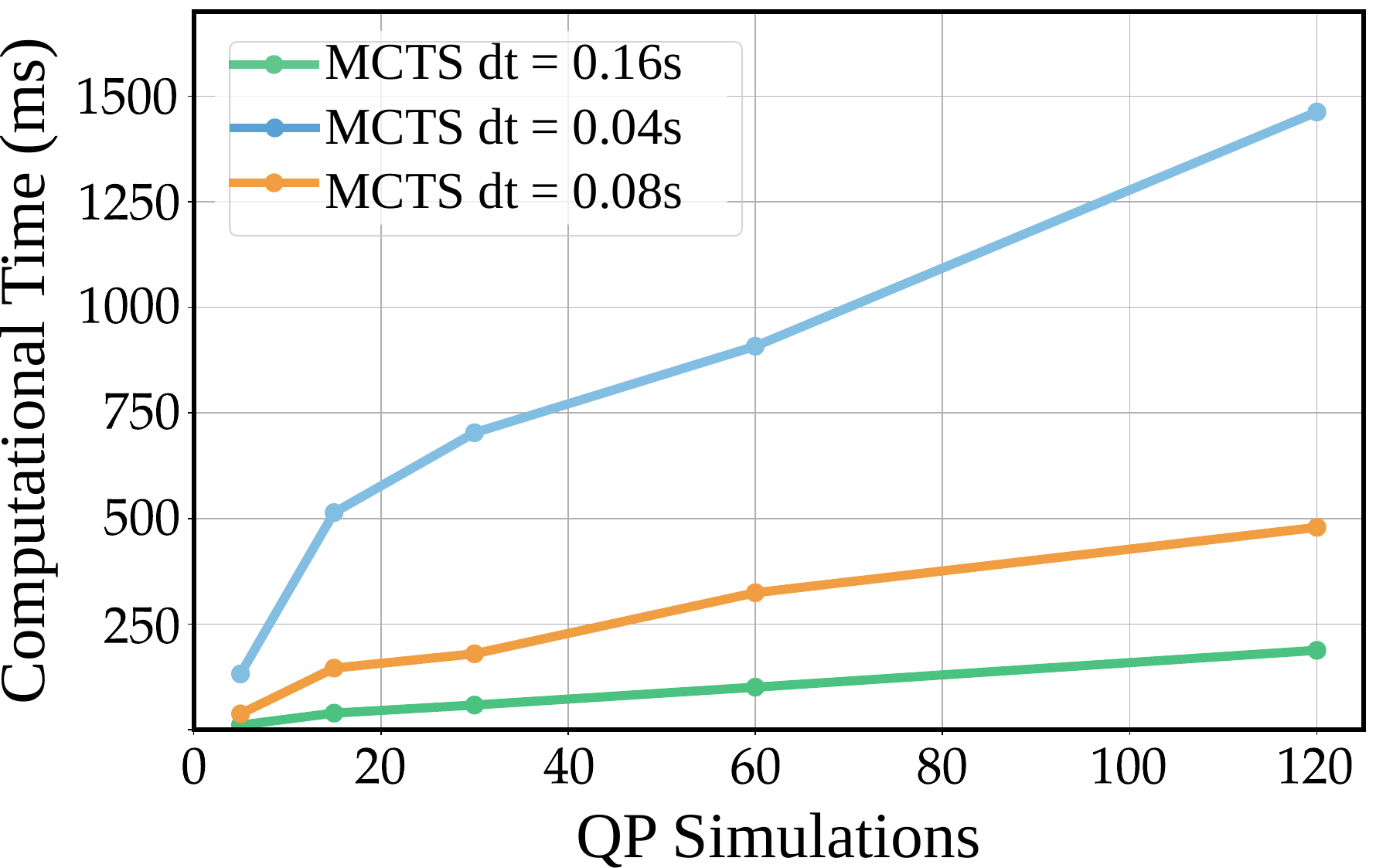}
    \caption{Comparison of the mean MCTS computation time for different tree discretizations and increasing number of simulations while disturbing the system along different swing phases with a force of $150$ N for $100$ ms.}
    \vspace{-15pt}
    \label{fig:mcts_computation_times_comparison}
\end{figure}

There are four main parameters that can affect the optimality of the planned gait sequence and the respective motion performance of the system. In this section, we present the quantitative evaluation of each parameter's influence on the MCTS solution used to determine our parameter choices. These are:

\begin{itemize}
  \setlength\itemsep{0.2em}
  \item \textit{Number of Simulations}: this is the number of times an incomplete gait sequence is randomly filled and the OCP described in Sec. \ref{sec:simulation} is solved.
  \item \textit{Tree Discretization}: the time resolution for each node's contact status, indicating the duration for which a specific contact status is maintained.
  \item \textit{Tree Horizon}: the length of the time horizon over which the gait sequence is optimized.
  \item \textit{Replanning Frequency}: the rate at which the gait sequence is updated per second.
\end{itemize}


To evaluate the parameters' importance, we compare their influence on the MPC tracking cost in simulation using the RaiSim physics engine~\cite{raisim}. The evaluations were conducted using the Aliengo robot model, a quadrupedal robot weighing $22$ kg and measuring $65$cm in length. In these evaluations, the robot is tasked with tracking a reference forward velocity of $0.3$ m/s while being laterally disturbed by a force of $150$ N for $100$ ms every $3$ s. The disturbances are applied $50$ times at five different instances of the swing phase ($0\%$, $20\%$, $40\%$, $60\%$, $80\%$). Unless stated otherwise, we assume a tree horizon value of $0.64$ s for the presented evaluations.

\subsection{Number of Simulations \& Tree Discretization}
\label{sec:simulations_discretization_comparison}

Figure~\ref{fig:mcts_real_cost_comparison} shows the MPC tracking cost for different MCTS tree discretization time $dt$ ($0.16$ s, $0.08$ s, $0.04$ s) and the number of MPC rollouts ($5$, $15$, $30$, $60$, $120$) performed at each node expansion. The tree discretization values are chosen to be multiples of $0.04$ s, which is the MPC discretization timestep, in order to ease the contact sequence conversion from the MCTS gait sequence to the MPC gait sequence.

Figure~\ref{fig:mcts_real_cost_comparison} highlights how large discretization time leads to very high MPC tracking costs, mainly due to the large model inaccuracies induced by the larger $dt$. Using a $dt$ that is twice as large as the MPC's discretization performs worse than using a $dt$ that matches the one of the MPC when the number of simulations is low. However, as the number of simulations increases, the performance of the two discretizations becomes almost identical. Figure~\ref{fig:mcts_real_cost_comparison} also presents evidence that a high number of simulations during the MCTS gait planning leads to better performance. This is due to a better cost estimate for the nodes as they are less sensitive to cost outliers brought by the random nature of the sampling process.

The number of simulations and the tree discretization not only affect the MPC tracking cost but also the computation time required for the algorithm to converge. Figure~\ref{fig:mcts_computation_times_comparison} shows how larger computation times are associated with a higher number of simulations and smaller tree discretizations. The higher number of MPC rollouts increases the convergence time due to the increasing number of QP problems that must be solved to evaluate the final cost associated with every node. The tree discretization time directly affects the MCTS depth, since the smaller the discretization time the higher the number of nodes that must be evaluated during the search.

Based on the MPC tracking cost shown in Fig. \ref{fig:mcts_real_cost_comparison}, we select $0.08$ s and $120$ as the best tree discretization time and number of MPC rollouts for the MCTS gait planning process, respectively. 
\subsection{Tree Horizon}
\label{sec:tree_horizon_comparison}

\begin{figure}[t]
    \centering
    \includegraphics[width=0.8\linewidth]{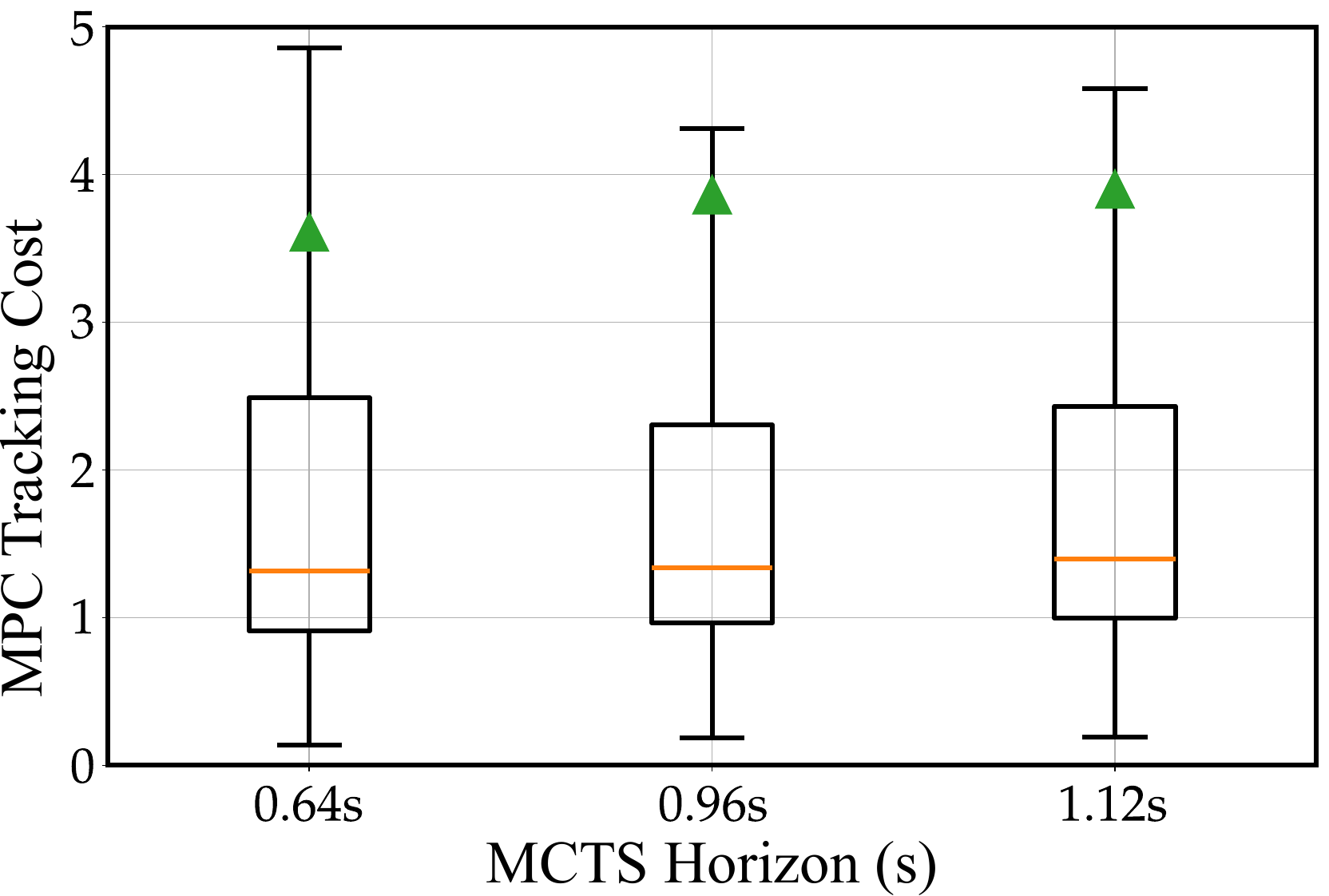}
    \caption{Comparison of the mean MPC tracking cost for a tree discretization $dt$ of $0.08$ s, $120$ MPC rollouts, and different tree horizons while disturbing the system along different swing phases with a force of $150$ N for $100$ ms.}
    \vspace{-10pt}
    \label{fig:mcts_horizon_comparison}
\end{figure}

Figure~\ref{fig:mcts_horizon_comparison} shows the influence of the tree horizon parameter on the MPC tracking cost when using a tree discretization of $0.08$ s and $120$ MPC rollouts. \revise{We can observe that longer horizons marginally change the cost, hinting at the fact that longer horizon plans are not crucial, as shown in \cite{CafeMpcAC}, as long as fast replanning is carried out (Sec.~\ref{sec:planning_frequency_comparison}). This conclusion may be task-dependent. In fact, while the reduced model’s approximation is dominant in the tested scenarios, for cases where the robot must execute long flight phases, such as sparse stepping stones, a longer horizon could become crucial.}

\subsection{Replanning Frequency}
\label{sec:planning_frequency_comparison}

As previously shown in Sec.~\ref{sec:simulations_discretization_comparison}, a high number of simulations is important to obtain a good cost estimate for the nodes. However, the higher the number of simulations the higher the time to complete the MCTS planning process. Therefore, it is important to establish the minimum replanning frequency that should be respected to maintain the best MPC tracking cost performance established in Sec.~\ref{sec:simulations_discretization_comparison}. To do so, we evaluate the influence that different replanning frequencies have on the MPC tracking cost. Since we evaluated the MCTS gait planner with frequencies as low as $2$ Hz, we need to make sure to have a long enough contact sequence to feed into the MPC. For this reason, we use a tree horizon of $1$ s.

\begin{figure}[t]
    \centering
    \includegraphics[width=0.8\linewidth]{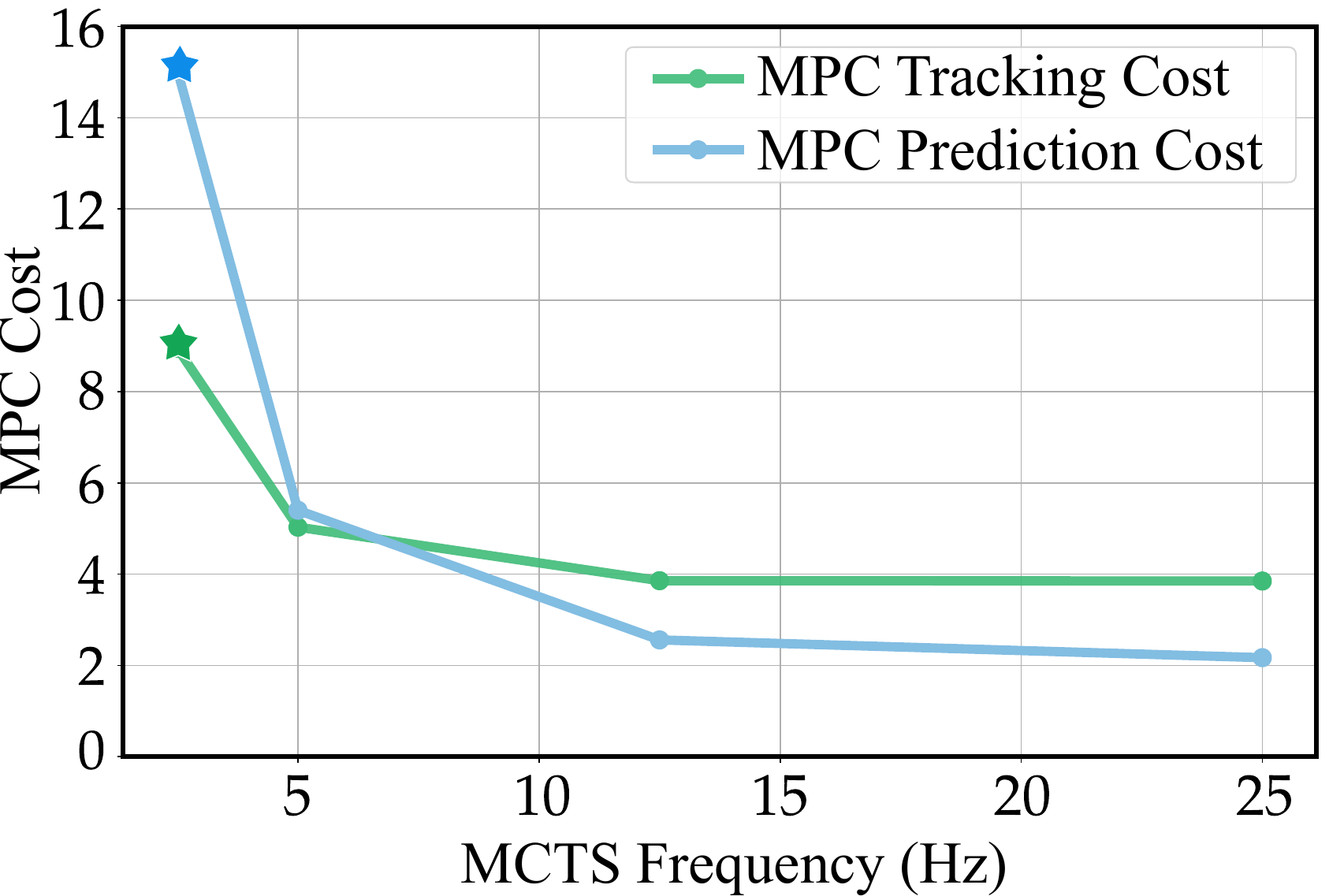}
    \caption{Comparison of the mean MPC tracking and prediction cost for a tree discretization $dt$ of $0.08$ s, $120$ MPC rollouts, and increasing MCTS planning frequencies while disturbing the system along different swing phases with a force of $150$ N for $100$ ms.}
    \vspace{-15pt}
    \label{fig:mcts_frequency_comparison_80ms}
\end{figure}

Figure~\ref{fig:mcts_frequency_comparison_80ms} shows the influence of the replanning frequencies on the MPC tracking cost and the MPC prediction cost. The figure shows that faster replanning improves the performance until $12.5$ Hz where it hits a plateau. Hence, in our setting, $12.5$ Hz is selected as the MCTS replanning frequency to be respected, which is also in line with a similar evaluation presented in \cite{meduri2023biconmp}. It should be noted that this update rate cannot be reached, with the currently available computational power, by the vanilla MCTS implementation described in Sec.~\ref{sec:vanilla_mcts}, whose real performance is indicated with the stars in Figure~\ref{fig:mcts_frequency_comparison_80ms}. Therefore, a significant speed-up is required to reach the established replanning frequency.

\section{MCTS Speed Up}
\label{sec:mcts_speedup}

In Sec.~\ref{sec:mcts_parametrization}, we presented the results of our ablation study on the MCTS parameters. From this study, as depicted in Fig. \ref{fig:mcts_frequency_comparison_80ms}, we identified a minimum replanning frequency of $12.5$ Hz to achieve good system performance. Additionally, we concluded that a greater number of simulations leads to a better MCTS solution and robot's performance. However, as shown in Fig. \ref{fig:mcts_computation_times_comparison}, increasing the number of simulations to better evaluate each expanded node also increases the computation time. This means that we can either allow the planner to perform more simulations by lowering the replanning rate, or maintain the minimum replanning frequency by limiting the MCTS to just $5$ rollouts per expansion. Both of these solutions, however, limit the performance of the vanilla MCTS, making its deployment on real hardware extremely challenging.

In this section, we present a simple yet effective way to overcome this issue. We propose a learning-based method to reduce the number of MPC rollouts to be solved, with the goal of making MCTS real-time on commonly available hardware without a substantial decrease in performance. \revise{We also describe our dataset generation process and the adopted architecture for the learning-based method.}

\subsection{\revise{Dataset \& Architecture}}
\label{sec:value_function_dataset_architecture}

\revise{The dataset for the training process is generated in simulation using the Raisim physics engine. We run the vanilla MCTS method with the best-identified parameters in Sec.~\ref{sec:mcts_parametrization}, which correspond to a dt of $0.08$ s and $120$ simulations. We trained on flat and rough terrains while randomizing the velocity commands, the step height, the robot's mass, the swing trajectory controller gains, and the robot's height. Additionally, we also perturb the system with randomly generated forces in terms of magnitude, duration, and frequency. We log each MCTS iteration input and output and use the nodes' cost estimates for training. This way, we collect a total of $26563$ MCTS trees from the simulations.}

\revise{A Multi-Layer Perceptron (MLP) with three hidden layers of $512$ neurons each, batch normalization layers, and dropout regularization are used as the underlying architecture. We use ReLU activation functions for the intermediate layers and a linear activation function for the output layer. The Adam optimizer~\cite{diederik2014adam} with a learning rate scheduler is used to optimize the network's weights.}

\subsection{Value Function Network}
\label{sec:value_function_network}

Given the computationally demanding cost of estimating expanded nodes by solving several MPC rollouts, we propose to learn a Value Function (VF) that approximates the cost-to-go $\bar{P}_n$ (Sec.~\ref{sec:simulation}). The network's input is defined as follows:
\vspace{-1pt}
\begin{equation*}
    \xv_{vf} = [\zv_{c, e}, \vv_{c,e},\Phim_e, \boldsymbol{\omega}^b_e, \rv, t_{swing}, t_{stance}, \cv_n] \in R^{78}
    \label{eq:value_function_input}
\end{equation*}

where $\zv_{c, e}$ is the error between the reference and actual CoM height, $\vv_{c, e}$ is the error between the reference and actual CoM linear velocity, $\Phim_e$ is the error between the reference and actual base orientation, and $\boldsymbol{\omega}^b_e$ is the error between the reference and actual base angular velocity. $\rv$ comprises the actual foot positions with respect to the CoM, $t_{swing}$ and $t_{stance}$ are the actual swing and stance time of each leg in seconds, and $\cv_n$ is the contact sequence that leads to the node $n$. If the node is not terminal, we fill the missing part of the contact sequence with $-1$ values, in order to make sure the input size to the network remains the same.

\revise{Estimating a node’s cost with the proposed VF takes, on average, less than 1 ms. This results in a substantial speed-up over the QP-based evaluation, since a comparable performance is only achieved with 120 simulations, which would take approximately 20 ms if running 10 processes in parallel.}

\subsection{Combined Approach}
Only relying on the learned value function to estimate $\bar{P}_n$ (\ref{eq:qps_cost}), can be detrimental when the states are out of the distribution of the training data.  This is a well-known problem of imitation learning \cite{ross2011reduction} from offline data, and it is likely to happen during the deployment of such networks in the real world. In this work, we seek to obtain generalization through the combination of model-based \revise{MPC rollouts} solutions with the bootstrapping obtained by employing the VF network.

Inspired by \cite{ross2011reduction}, we perform a simple update rule for the node cost, such as
\begin{equation*}
    P_n = \alpha\bar{P}_n + (1-\alpha)\bar{P}_{vf}
\end{equation*}
with $\alpha$ being a heuristically chosen parameter, \revise{$0.75$ in our case}. Note that, contrary to \cite{ross2011reduction}, we keep $\alpha$ fixed to allow the robot to react in new situations.
As we will see in the result presented in Sec.~\ref{sec:results}, $\alpha$  makes a trade-off between trusting the learned VF and online MPC rollouts.

\section{Results}
\label{sec:results}

In this section, we present the evaluation of our proposed method in simulation using the RaiSim physics engine and on a real electric quadruped.

\revise{We perform several evaluations of our proposed method in simulation. We first compare a vanilla MCTS gait planner against two purely learning-based methods and our proposed hybrid method,  using the same evaluation method described in Sec.~\ref{sec:mcts_parametrization}. Then, we present a quantitative analysis on the impact of MPC rollouts on the proposed hybrid approach in an extreme out-of-distribution (EOD) situation. Finally, we conduct a comparison between our proposed hybrid method against trotting gait sequences that assume periodic contact timings with different step frequencies.}

On hardware, using a real electric quadruped, we demonstrate the pipeline's disturbance rejection performance in comparison to a fixed periodic gait. As highlighted in the accompanying video, this is the first successful real-time implementation of a sampling-based method for non-gaited locomotion. Moreover, we perform a qualitative comparison between simulation and hardware results.

For both simulation and hardware evaluations, we set a maximum allowed time budget of $80$ ms for the MCTS gait planner in order to meet the $12.5$ Hz replanning frequency identified in Sec.~\ref{sec:mcts_parametrization}.

\subsection{Simulation results}
\label{sec:simulation_results}

\subsubsection{Baseline Comparison}
\label{sec:baseline_comparison}

\begin{figure}[t]
    \centering
    \includegraphics[width=0.9\linewidth]{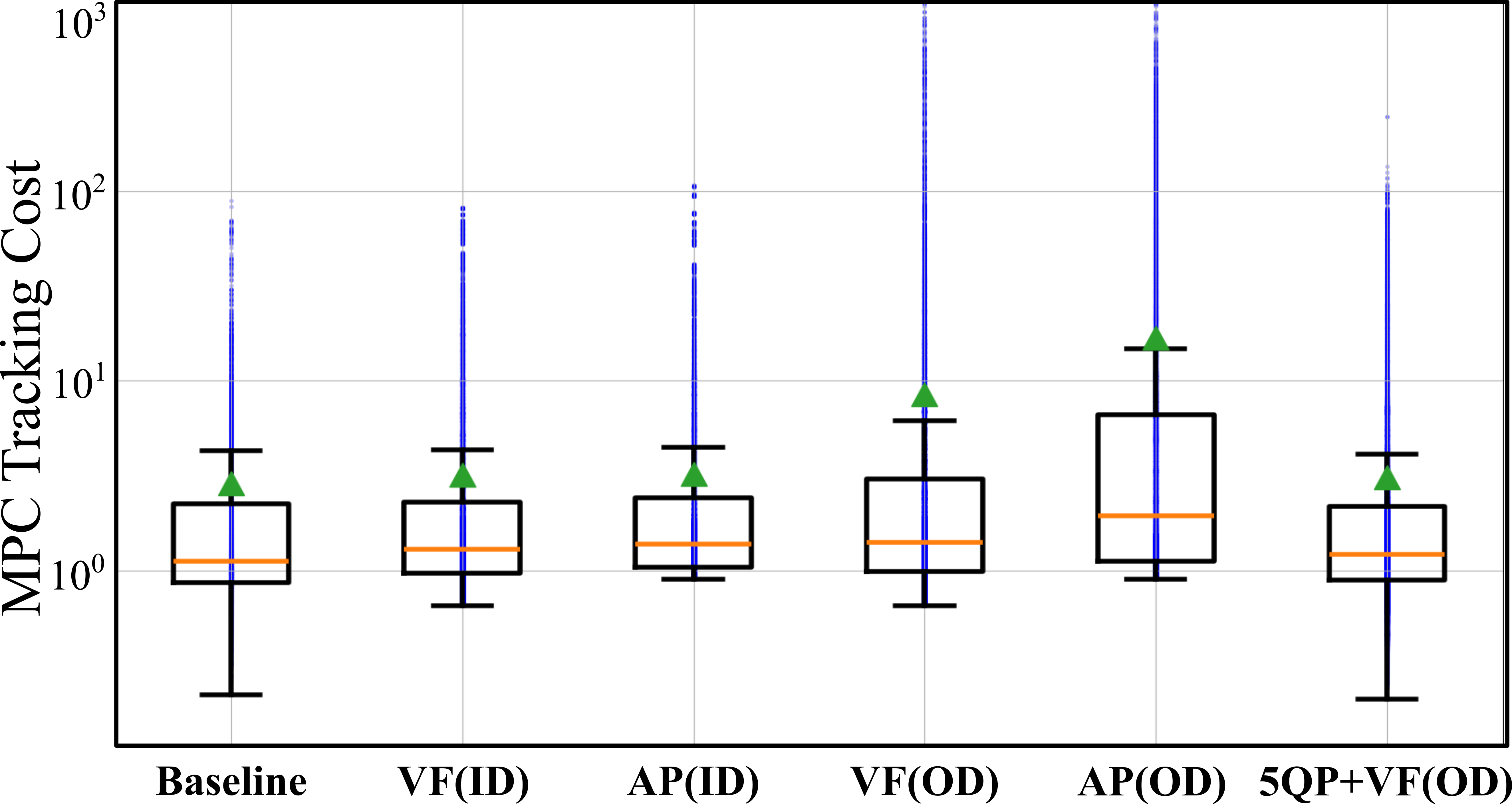}
    \caption{Comparison of how the MPC tracking cost distribution varies, inside (ID) and outside (OD) the learning distribution for different approaches. With a green triangle, we showcase the mean MPC tracking cost. From left to right we show the comparison between the MCTS Baseline against four purely learning-based approaches based on VF and AP approaches for both ID and OD scenarios, and a hybrid approach performing additional model-based MPC rollouts with VF bootstrapping in an OD scenario. The blue points represent the MPC tracking error samples for each approach. All proposed approaches see an increase in the mean MPC tracking cost compared to the baseline of $11\%$, $12\%$, $195\%$, $484\%$, and $7\%$ respectively.}
    \vspace{-15pt}
    \label{fig:mcts_vf_approaches_comparison}
\end{figure}

we evaluate the following three approaches against a vanilla MCTS gait planner baseline:

\vspace{0.5em}

\begin{itemize}
  \item VF
  \item Action Policy (AP)
  \item Combined Approach (5 QP + VF)
\end{itemize}

\vspace{0.5em}

The first two methods are purely learning-based while the third one is a hybrid approach that combines the learned VF network with $5$ model-based MPC rollouts. The AP outputs the next optimal contact directly, thereby obtaining the full sequence by successively querying the network. We add such a method in the analysis to provide a complete insight into the benefit of our proposed method. Both VF and AP share the same network architecture, a Multi-Layer Perceptron (MLP) with three hidden layers of $512$ neurons each.

We carry out a comparison, always in terms of MPC tracking cost, on two different scenarios: Inside Distribution (ID) and Outside Distribution (OD). In the ID case, we train two different models for various target speeds while including, in the training data, external disturbances to the robot base in the form of pushes. On the other hand, in the OD case, we only train the models with different target speeds without taking into consideration any disturbance force. Figure~\ref{fig:mcts_vf_approaches_comparison} shows the results for both scenarios. 

In the ID scenario, both the VF and AP learning-based models perform on par with the baseline in terms of mean and variance of the MPC tracking cost. The mean MPC tracking cost is almost identical between the three approaches, showing a good overall approximation by the learning-based methods. The MPC tracking cost distribution for both VF and AP tends to reach higher peaks compared to the baseline, although being within close range. Overall, for both purely learning-based approaches, if they are trained on a diverse enough dataset that covers the state space of the system, they show similar performance as the baseline.

In the OD scenario, the learning-based models undergo a substantial increase in the mean MPC tracking cost compared to the baseline, with the VF showing a better performance compared to AP. This is primarily because, in the case of the VF, constraints are imposed during the expansion process, whereas for the AP, they are integrated as part of the learned behavior.

Observing the results for the third comparison, we see that the combination of the VF with only $5$ MPC rollouts significantly lowers the mean MPC tracking cost and brings back its distribution to a range similar to the baseline. This demonstrates the benefit of our proposed hybrid approach in OD cases, where a few QP model-based simulations can help maintain a certain level of robustness and performance while being capable of running at the desired replanning frequency. 

\subsubsection{\revise{MPC Rollouts Impact on the Hybrid Approach}}
\label{sec:mpc_rollouts_comparison_eod}

\begin{figure}[t]
    \centering
    \includegraphics[width=0.9\linewidth]{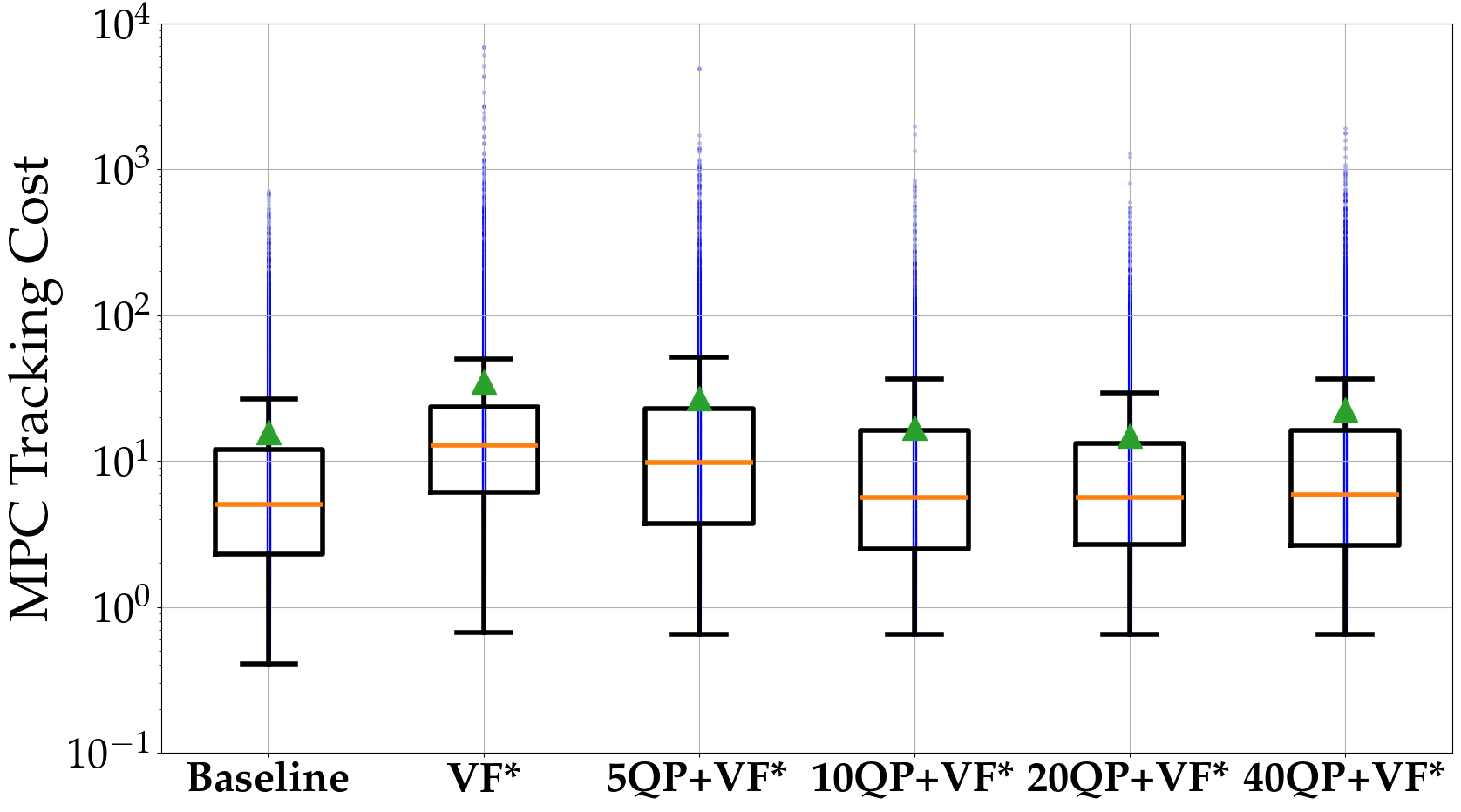}
    \caption{\revise{Comparison of how the MPC tracking cost distribution varies for an extreme outside of distribution (EOD) scenario. From left to right we show the comparison between the MCTS Baseline, using a tree discretization of $0.08$ s and $120$ simulations, against a pure VF learning-based approach, and four hybrid approaches performing respectively $5$, $10$, $20$, and $40$ additional model-based MPC rollouts with VF bootstrapping. The blue points represent the MPC tracking error samples for each approach. With the * we mean a VF trained according to the description in Sec.~\ref{sec:mpc_rollouts_comparison_eod}. All proposed approaches see an increase in the mean MPC tracking cost compared to the baseline of $123\%$, $73\%$, $7\%$, $5\%$, and $44\%$ respectively.}}
    \vspace{-15pt}
    \label{fig:mcts_eod_comparison}
\end{figure}

\revise{Figure~\ref{fig:mcts_eod_comparison} shows the effect of the numbers of MPC rollouts on our proposed hybrid approach in an EOD scenario. We compare the hybrid approaches with an MCTS vanilla baseline and a pure VF learning-based approach. The EOD scenario is set such that the VF is trained only on data containing forward velocities up to $0.5$ m/s on a flat terrain, but we test on a sloped terrain with a commanded forward velocity of $0.6$ m/s without exerting external disturbance forces.}

\revise{The results, similar to Sec.~\ref{sec:baseline_comparison}, show how a pure VF approach performs worse in OD scenarios compared to a hybrid approach, where $5$ MPC rollouts are performed. However, we also note that increasing the number of MPC rollouts performed in the hybrid approach does not necessarily lead to better system performance. Since we enforce an optimal replanning frequency of $12.5$ Hz for all configurations, as found in Sec.\ref{sec:planning_frequency_comparison}, the hybrid MCTS has limited time available to return a solution. This means that by increasing the number of simulations per node expansion, the MCTS spends the majority of its time estimating fewer node costs with MPC rollouts, leaving the rest to be estimated using VF alone.
Given our computational resources, the average percentage of nodes evaluated with MPC rollouts decreases from $39\%$ with $5$ rollouts per node to $27\%$ for $40$ rollouts. Consequently, the mean MPC tracking cost improvement stagnates, showing only minor gains when increasing the number of rollouts up to 20, while system performance actually degrades with 40 rollouts.}

\revise{In conclusion, increasing the number of MPC rollouts in the hybrid MCTS is not always advantageous. A balance must be found between using more rollouts per node to obtain fewer but more accurate QP-based cost estimates or reducing the number of rollouts to estimate less accurate costs across a larger number of nodes.}

\subsubsection{Comparison with Periodic Contact Scheduling}
\label{sec:periodic_contacts_comparison}

\begin{figure}[t]
    \centering
    \includegraphics[width=0.9\linewidth]{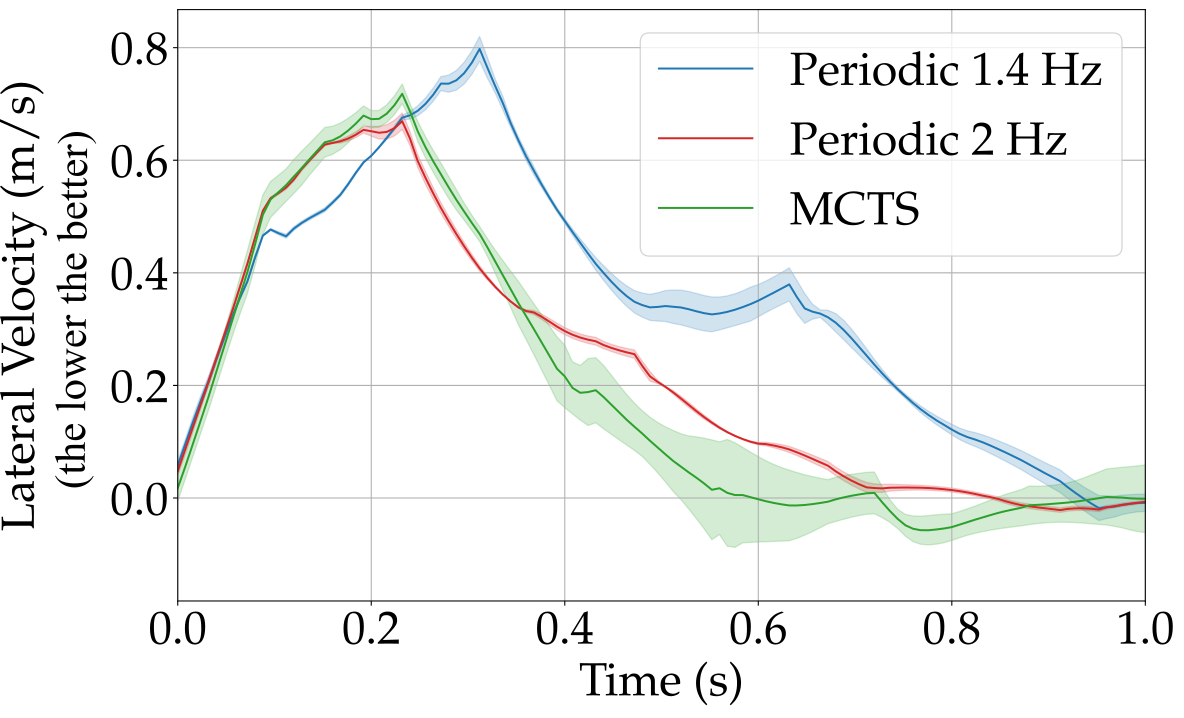}
    \vspace{-10pt}
    \caption{\revise{Comparison of the lateral velocity (the solid colors represent the mean value and the shaded areas the confidence band) between two periodic trot gait sequences with a respective step frequency of $1.4$ Hz and $2$ Hz against our proposed MCTS gait planner. During walking, we apply a disturbance at $t = 0$ s with a positive force of $150$ N for $100$ ms in the robot's lateral direction on a sloped terrain. We repeat the disturbance process $50$ times for each method. As shown in the plot, the MCTS results in an overall lower lateral velocity after disturbance.}}
    \vspace{-6pt}
    \label{fig:mcts_mpc_comparison}
\end{figure}

Figure \ref{fig:mcts_mpc_comparison} presents a comparison of the \revise{mean lateral velocity and respective confidence band} between a non-gaited sequence, obtained using MCTS, and two periodic trotting gaits: one with a relatively low stepping frequency of $1.4$ Hz and the other with a higher stepping frequency of $2$ Hz. For both frequencies, we consider a step duty factor of 0.6 (ratio between the stance period and the whole step cycle period). \revise{The values shown in Fig.~\ref{fig:mcts_mpc_comparison} are obtained by pushing the system $50$ times at time $0$ s with a force of $150$ N in the robot's lateral direction for $100$ ms while tracking a forward velocity of $0.3$ m/s.}

\revise{Our method and the $2$ Hz trotting gait show a similar overall performance. They both peak in terms of recorded lateral velocity after $0.2$ s from the moment of the push, with our method having a slightly higher peak lateral velocity of $0.72$ m/s compared to the $0.63$ m/s for the $2$ Hz trotting gait. On the other hand, the $1.4$ Hz trotting gait peaks only at $0.3$ s with a peak lateral velocity of $0.8$ m/s.}

\revise{In general, we also note how our proposed method is able to bring the system back to zero lateral velocity at around $0.6$ s which is significantly faster than the $0.72$ s for the $2$ Hz trotting gait and the $0.96$ s for the slower $1.4$ Hz trotting gait. The superior performance of our method comes from its ability to optimize both timing and gait sequence, enabling it to outperform conventional periodic gaits.}

\subsection{Hardware Experiments}
\label{sec:hardware_experiments}

\begin{figure}[!t]
    \centering
    \includegraphics[width=1.0\linewidth]{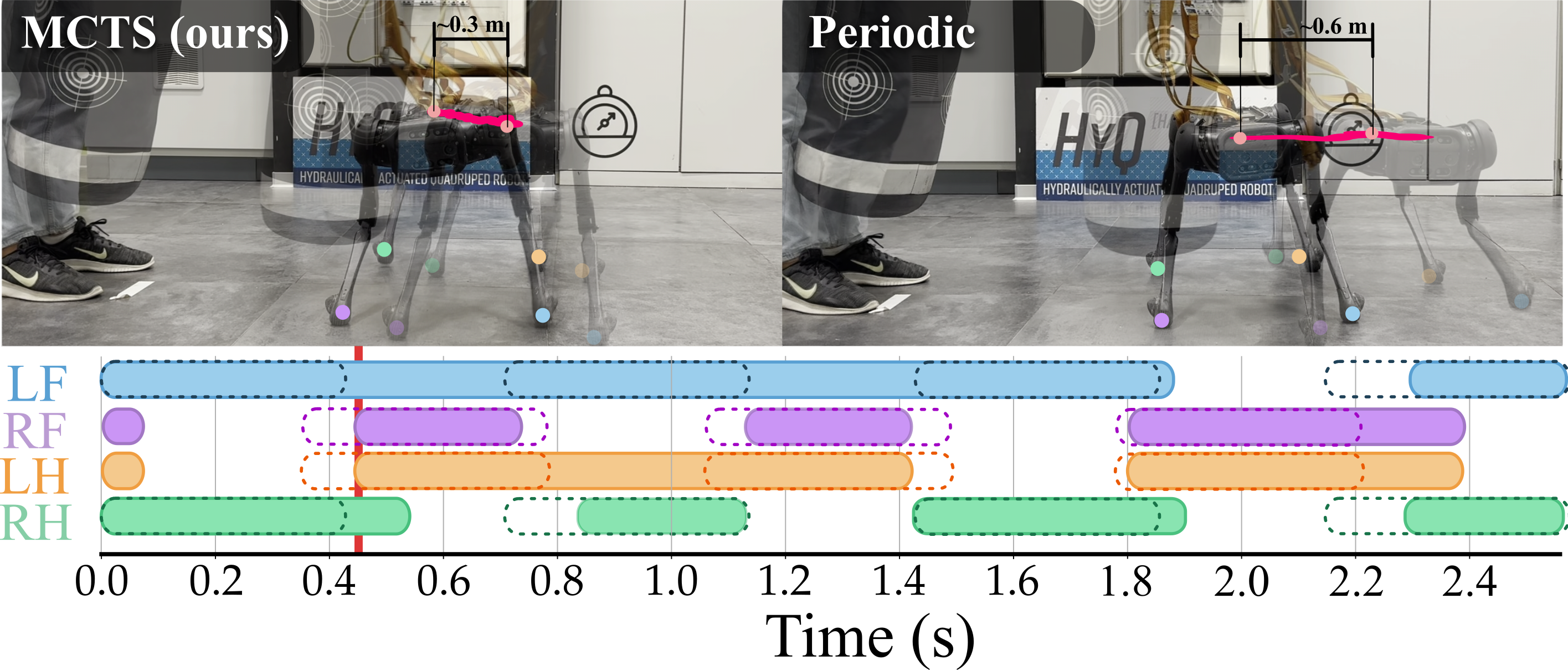}
    \vspace{-19pt}
    \caption{On top, snapshots of the experiments on the Aliengo platform. On the left, the robot used the presented MCTS gait planner. On the right, a periodic trot with the same reference step frequency used by MCTS was used. A punching bag let go from the same height pushes the robots in a repeatable manner in both cases. In pink is the trace of the robot trunk, the smaller the traveled distance, the  better the disturbance rejection. On the bottom, are the plots of the contact sequence chosen by the MCTS (solid color) and the periodic gait (dashed line) for each leg, where we highlight in red the moment of contact between the robot and the punching bag.}
    \vspace{-15pt}
    \label{fig:exp}
\end{figure}

To validate the proposed method, we also performed tests on real hardware. Thanks to the speed-up achieved by our hybrid approach, we are able to reach the necessary replanning rates for deployment on a real quadruped. We test the pipeline on Aliengo, an electric quadruped robot made by Unitree Robotics \cite{aliengo} that weighs around 22 kg. The entire pipeline runs externally on a 12th-generation Intel i7 processor. In Fig. \ref{fig:exp}, we present some snapshots of the experiments while also highlighting the contact sequence generated by the proposed MCTS gait planner. To ensure the repeatability of the scenario, the robot is pushed by a 27 kg punching bag hanging from a crane. Our method, exploiting the MCTS gait adaptation, is compared with a periodic approach where the robot trots at a fixed frequency of $1.4$ Hz and duty factor 0.6.

As shown in the diagram, the MCTS gait planner adapts the contact sequence by keeping both left feet on the ground after the impact to better counteract the push before returning to the more efficient trotting frequency incentivized by the cost in~\eqref{eq:qps_cost}. The improved disturbance rejection can be observed by the trajectory of the trunk's position (illustrated by a pink line). These results and further tests on the robot can be seen in the accompanying video.

\section{Conclusions}
\label{sec:conclusions}

In this work, we presented a novel approach for non-gaited legged locomotion that extended the work in \cite{amatucci_mcts} by bringing to the framework real-time capability and the first-ever successful implementation on hardware of such a sampling approach.
We offered an extensive analysis of the parametrization of the MCTS formulation for non-gaited locomotion and compared it to standard control approaches that assume periodicity in the gait sequence, ultimately showcasing the benefit of our approach over such methods.

Future works will focus on integrating visual feedback into our formulation combining it with a surface selection method such as \cite{dhedin2024diffusion}. Furthermore, we aim to incorporate a more complex but efficient robot model like the one used in \cite{amatucci2024} to enable robust locomotion for multi-legged systems such as quadrupeds and humanoids.

\bibliographystyle{IEEEtran}
\bibliography{bibliography}
\end{document}